\documentclass{IEEEtran}

\usepackage{cite}

\usepackage[pdftex]{graphicx}
\DeclareGraphicsExtensions{.pdf,.jpeg,.png}

\usepackage[cmex10]{amsmath}
\usepackage{amssymb}

\usepackage{mdwmath}
\usepackage{mdwtab}

\usepackage[caption=false,font=footnotesize]{subfig}

\usepackage{url}

\usepackage[utf8]{inputenc}
\usepackage{cleveref}
\usepackage{threeparttable}
\usepackage{color}  
\usepackage{soul}

\usepackage{enumitem}
\setenumerate[1]{label=(\arabic*)}

\def\eg{\emph{e.g., }}
\def\ie{\emph{i.e., }}
\def\wrt{\emph{w.r.t. }}

\begin{document}

\title{A dataset for benchmarking vision-based localization at intersections\\
~\\
\small{supplemental material for ``Vision-Based Localization at Intersections using Digital Maps''}}

\author{Augusto L. Ballardini, Daniele Cattaneo, and Domenico G. Sorrenti%
\thanks{Dept. Informatica, Sistemistica e Comunicazione, Universit\`a degli Studi di Milano - Bicocca, Milano, Italy; email: augusto.ballardini@unimib.it, daniele.cattaneo@disco.unimib.it, domenico.sorrenti@unimib.it}}

\maketitle

\begin{abstract}
In this report we present the work performed in order to build a dataset for benchmarking vision-based localization at intersections, \ie a set of stereo video sequences taken from a road vehicle that is approaching an intersection, altogether with a reliable measure of the observer position. This report is meant to complement our paper ``Vision-Based Localization at Intersections using Digital Maps'' submitted to ICRA2019. It complements the paper because the paper uses the dataset, but it had no space for describing the work done to obtain it. Moreover, the dataset is of interest for all those tackling the task of online localization at intersections for road vehicles, \eg for a quantitative comparison with the proposal in our submitted paper, and it is therefore appropriate to put the dataset description in a separate report. We considered all datasets from road vehicles that we could find as for the end of August 2018. After our evaluation, we kept only sub-sequences from the KITTI dataset. In the future we will increase the collection of sequences with data from our vehicle.
\end{abstract}


\section{Introduction}
In order to benchmark vision-based localization at intersections we need at least the following streams of data from a vehicle approaching an intersections: the streams from the two cameras of a stereo rig, and a reasonably accurate position estimate of the vehicle, to be used as GT (Ground Truth). Quite un-expectedly, given the number of datasets from road vehicles available today, collecting such sequences resulted very difficult. We analyzed the sequences of the many available datasets, and checked whether everything needed was usable. Unfortunately, it turned out that, for various reasons, most of the material publicly available was not usable. This report describes how we ended up with a few more than forty usable sequences of a vehicle approaching an intersections, all from the KITTI dataset residential sequences. Although all sequences are from the same German city, which could compromise the generality of the dataset \wrt the variety of the world-wide intersections, the dataset does include different intersection geometries, (slightly) different lightning, and different traffic conditions. In spite of the effort put in building the dataset, and despite its uniqueness, we believe this dataset should be integrated with sequences from other countries, and more extreme light and traffic conditions. Nevertheless, as for today, we believe it is the best mean to benchmark a proposal about vision-based vehicle localization at intersections.

We analyzed the following datasets: KITTI \cite{Geiger2013IJRR}, MALAGA \cite{malaga_blanco2013mlgdataset}, Oxford RobotCar \cite{oxford_RobotCarDatasetIJRR}, Rawseeds \cite{rawseeds}, and ApolloScapeAuto \cite{apolloscape_arXiv_2018}.

\begin{itemize}
  \item KITTI: used, a stereo rig and GPS-RTK are available;
  \item Malaga: unused for the lack of GPS-RTK;
  \item Oxford RobotCar: unused for the lack of GPS-RTK;
  \item Rawseeds: no intersection between real roads is available, only intersections between roads in private areas, whose resemblance to real intersections has been considered not good enough;
  \item ApolloScapeAuto: unused for the lack of GPS-RTK.
\end{itemize}

This report describes the following aspects, which are relevant parts of the work performed on the KITTI dataset in order to extract the largest set of sequences of approaches to intersections: the intersection model is introduced in \Cref{sec:intersectionmodel}, the issues with the camera calibration are reported in \Cref{sec:cameracalibration}, the issues with the alignments in OpenStreetMap are reported in \Cref{sec:openstreetmapalignemnt}, results about the sequences ending into the dataset are reported in \Cref{sec:results}. A short conclusion is then followed by the aerial view of the KITTI residential sequences, in \Cref{sec:appendixkittiresidentialsequences}.

\section{Intersection Model}\label{sec:intersectionmodel}
\begin{figure}
  \begin{center}
  \includegraphics[width=.48\textwidth]{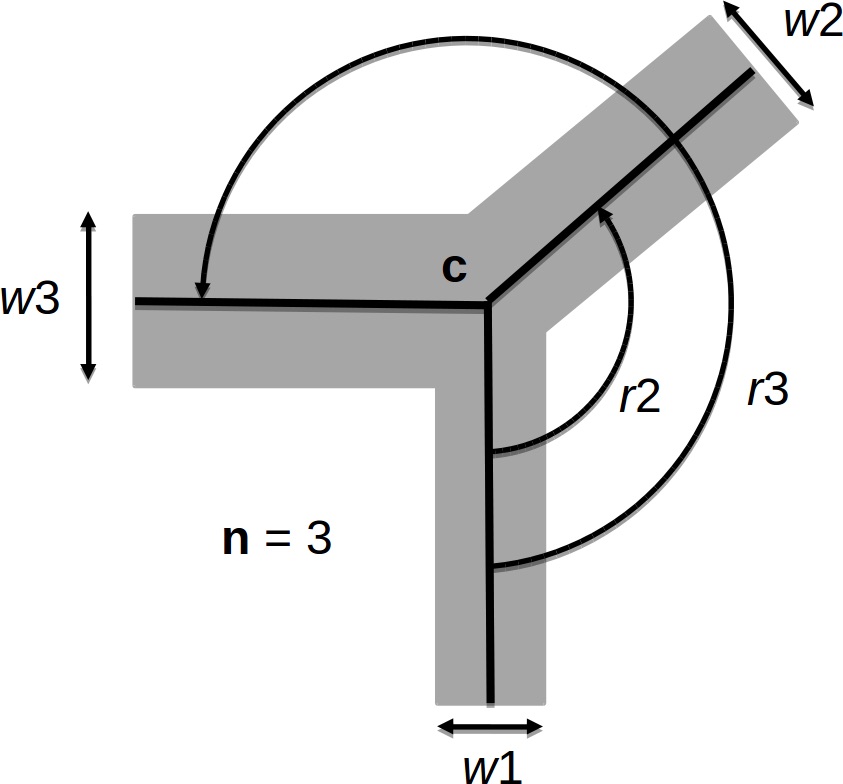}
  \end{center}
  \caption{The intersection model used to generate the intersections: $c$ represents the intersection center, $n$ the number of road segments, $w_i$ and $r_i$ the width and the orientation \wrt the road segment where the vehicle is.}
  \label{fig:modelgeometry}
\end{figure}

The configuration of the road arms coming to an intersection bases on an intersection model, proposed in \cite{Ballardini2017}, which is an enhancement of the one presented in \cite{Geiger2011b}, \cite{Geiger2014}. It is depicted in \Cref{fig:modelgeometry}, and has the following parameters:
\begin{itemize}
  \item The distance $ \textbf{c} $ of the intersection center, \wrt the hypothesized vehicle position;
  \item The number $ \textbf{n} $ of road segments / arms involved in the intersection;
  \item For each road segment $i$, the width $w_i$ and its orientation $r_i$ \wrt the road segment where the vehicle is.
\end{itemize}

\begin{figure}
  \begin{center} \includegraphics[width=0.48\textwidth]{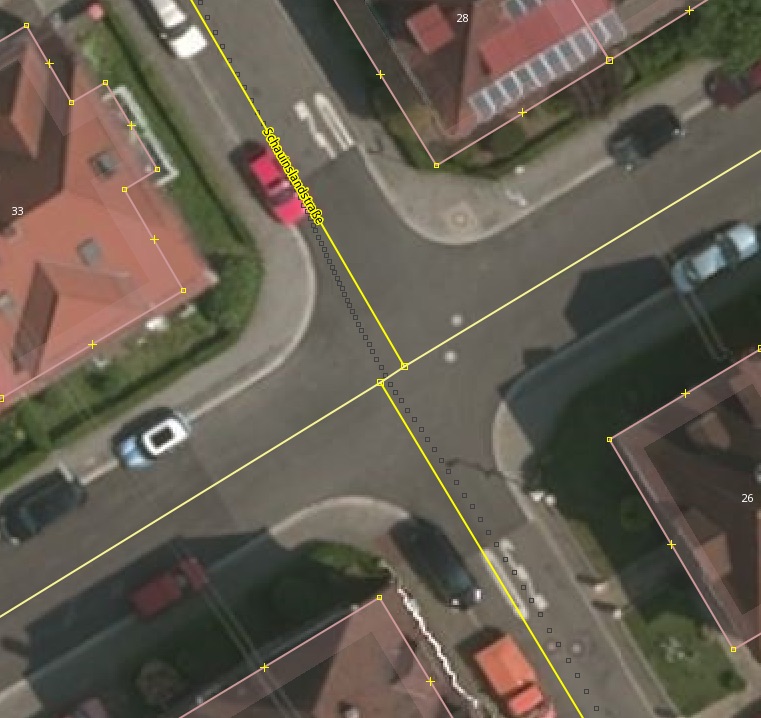} \end{center}
  \caption{Intersections that can not be represented using the model proposed in \Cref{fig:modelgeometry}: arm laterally displaced \wrt one another.}
  \label{fig:cannotfit1}
\end{figure}

This model allows us to represent almost every type of simple intersection, except the ones that can be better represented by roundabouts. A known limitation is the case of road segments entering the intersection with a lateral offset, as this offset is not available in the model. This makes it impossible to accurately represent conditions like the one in \Cref{fig:cannotfit1}.

\begin{figure}
  \begin{center} \includegraphics[width=0.48\textwidth]{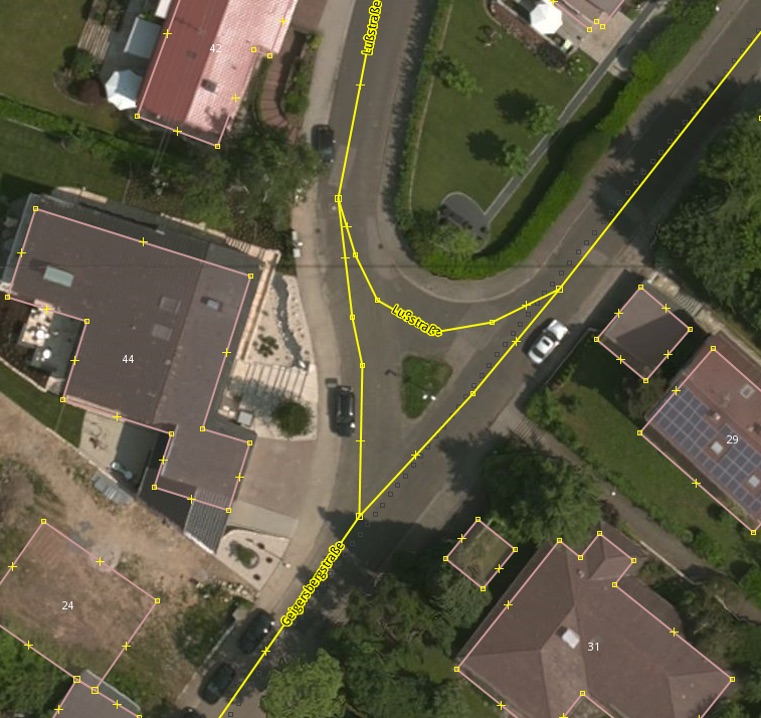} \end{center}
  \caption{Intersections that can not be represented using the model proposed in \Cref{fig:modelgeometry}: traffic island or other raised areas.}
  \label{fig:cannotfit3}
\end{figure}

\begin{figure}
  \begin{center} \includegraphics[width=.49\textwidth]{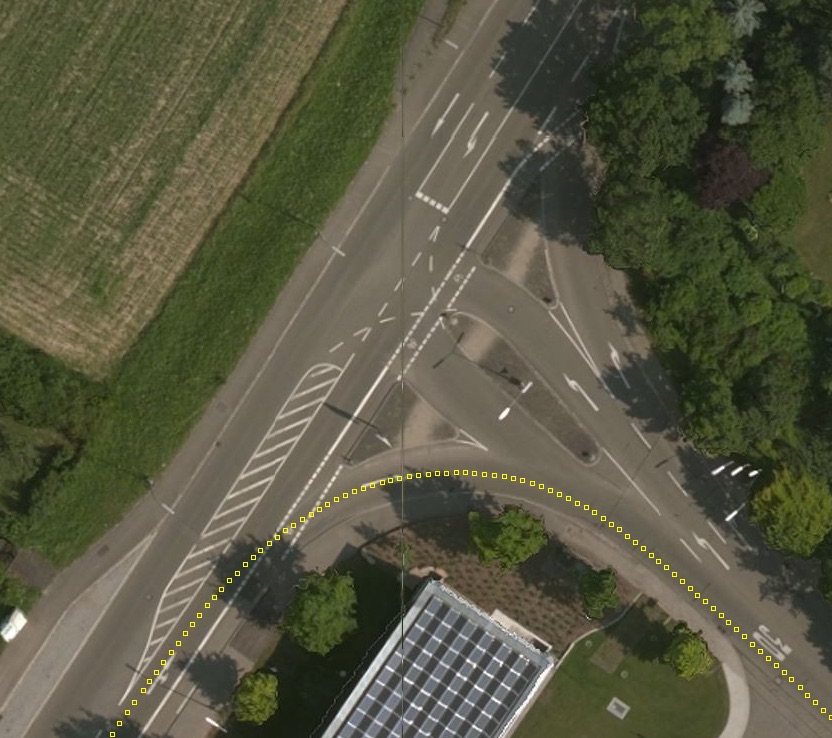} \end{center}
  \caption{The figure depicts intersections that cannot be represented using the model proposed in \Cref{fig:modelgeometry}: multiple laterally displaced arms plus traffic island or other raised areas; the yellow dotted line does not carry any particular meaning, please just dis-regard it.}
  \label{fig:cannotfit6}
\end{figure}

Further limitations arises with less frequent, but still existing configurations, \eg containing traffic islands or other raised areas as in \Cref{fig:cannotfit3}.
Other less frequent limitations concern intersection topologies that are available in the KITTI residential sequences, but cannot be dealt with the proposed model. As an example, see
\Cref{fig:cannotfit6}. The frames associated with these intersections are not included in the dataset.

\section{Camera Calibration}\label{sec:cameracalibration}

\begin{figure}
  \begin{center} \includegraphics[width=0.49\textwidth]{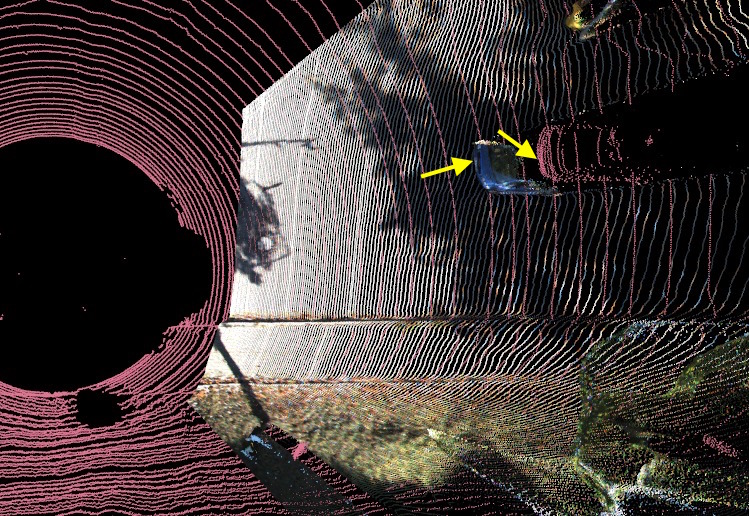} \end{center}
  \caption{Misalignment between LIDAR and camera-based PCs. The camera-based PC has been computed using the colour camera images and the projection parameters obtained for the colour cameras, while the other PC is from the LIDAR. The two has been superimposed using the poses of the two sensors (the stereo rig and the LIDAR) \wrt the vehicle. At the car bumper the misalignment is just more than 1.9m.}
  \label{fig:lasercameradisallineamento}
\end{figure}

Camera calibration has been found being the source of a significant problem. When computing the Point Cloud (PC) from the images of the colour cameras, we have been obtaining reconstructions of the intersections displaced from the correct positions. In order to find out where the problem was, we superimposed the LIDAR PC to the camera-based PC, by exploiting the poses of the two sensors \wrt the vehicle, an example can be seen in \Cref{fig:lasercameradisallineamento}. The figure clearly shows the different reconstruction produced by the camera processing pipeline. It can be noticed that the errors seem also to depend on the distance, so the error is not just a translation.

Thinking that a mistake could have been happened while putting online the dataset, we firstly re-computed the projection parameters of the colour cameras, using the online KITTI camera calibration tool as well as the KITTI calibration images. Unfortunately, this did not solve the problem.

We then discovered that these errors were not present for the reconstructions produced by the processing pipeline used in a previous paper of ours \cite{Ballardini2017}. The work in this paper, which was dealing with the problem of recognition of the topology of the intersection, was actually computing the 3D reconstruction from the black-and-white cameras of the vehicle, by exploiting the SGBM \cite{sgbm_Hirschmuller2008} stereo approach.

\begin{figure}
  \begin{center} \includegraphics[width=0.49\textwidth]{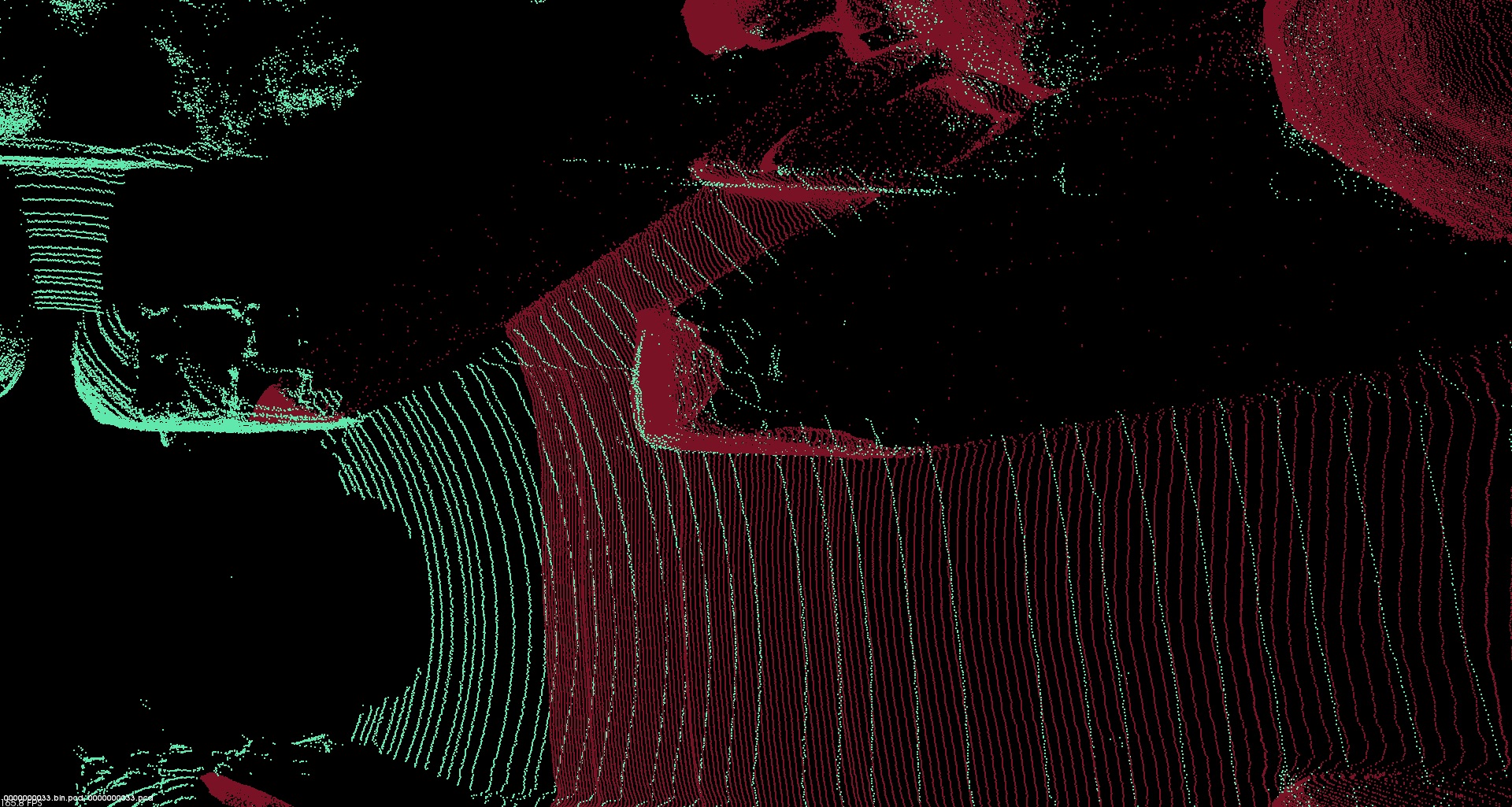} \end{center}
  \caption{Misalignment between LIDAR and camera-based PCs. The camera-based PC has been computed using the colour camera images and the projection parameters obtained for the black-and-white cameras, while the other PC is from the LIDAR. The two has been superimposed using the poses of the two sensors (the stereo rig and the LIDAR) \wrt the vehicle. Here the misalignment is less than 0.2m.}
  \label{fig:lasercameradisallineamento_ok}
\end{figure}

As a matter of coincidence, we then discovered that the - apparently non-sensical - substitution of the projection parameters of the colour cameras with the projection parameters of the black-and-white cameras, resulted in a reconstructed PC that was more accurately aligned with the LIDAR PC, see \Cref{fig:lasercameradisallineamento_ok}.

Given the time pressure, we gave up to discovering the reasons for the misalignment between the PCs, and proceeded using the projection parameters estimated for the  black-and-white cameras by the KITTI team as if they were the projection parameters of the colour cameras.

\section{OpenStreetMap Alignment}\label{sec:openstreetmapalignemnt}
Unfortunately, we also had to deal with alignment issues of the map. OpenStreetMap has some global map alignment problem, \ie some parts of map are not aligned to each other. The problem manifests with the GT trajectory of the vehicle being apparently out of the road, \eg on curbs or inside private areas. This is a known issue and there are also literature contributions on how to increase the quality of the alignments. There are different approaches dealing with this issue, see \eg \cite{MattyusRoadMaps,MattyusHDMaps,Mattyus_2017_ICCV,Wang_2017_ICCV}. Notice that there is no clear way to determine whether the culprit is the GT or the map. 

We started by superimposing the LIDAR PCs onto the aerial view, by means of the position GT, which is available in KITTI for each LIDAR PC. Of course, the misalignments between, \eg the boundaries of the buildings in the aerial view \wrt the same into the LIDAR PC cannot safely be attributed to an error in the pose GT or in the aerial map. This superimposition has been performed both with JOSM (the OpenStreetMap editor), requiring a cumbersome manual image mosaicing, and also with MAPViz, a ROS tool for superimposing layers of geo-localized data. Unfortunately, in both cases we could not obtain geometrically consistent results.

A further option is to use the capability of JOSM to set some alignment points in the aerial images, while the global position of the alignment points could be retrieved from a specific server. At this point we could modify the GT, by checking the alignment of buildings and roads. This would have been a difficult and very cumbersome task.

Alternatively, we could join, in a SLAM-like fashion, all aerial images, and then draw the roads and the GT position into this map (and re-entry all such data into OpenStreetMap), again a difficult and very cumbersome task.

In the end, we decided not to try to increase the quality of the maps, neither to correct the GT. Instead, by exploiting the spatial locality of the problem (\ie localization while approaching an intersection), we selected only the sequences taking place in areas where the alignment was good enough, according to the criteria described below. This has been a much simpler task. The criteria to select a sequence is as follows.

\begin{figure}
  \begin{small}
    \hspace{0.35cm} EXPECTEDOG \hspace{0.8cm} SENSOROG \hspace{1.0cm} OVERLAY
    \vspace{-0.15cm}
  \end{small}
  \begin{center}     
    \subfloat[Example of correct alignment]{\includegraphics[width=0.99\columnwidth]{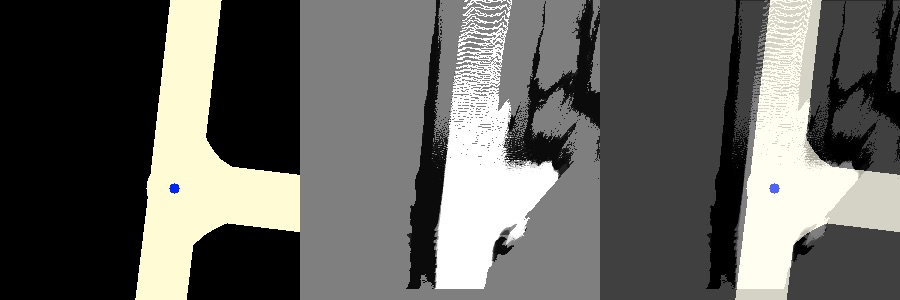}\label{fig:2679_gt_pcr_overlay}}\\ 
  \end{center}
  \begin{small}
    \hspace{0.35cm} EXPECTEDOG \hspace{0.8cm} SENSOROG \hspace{1.0cm} OVERLAY
    \vspace{-0.15cm}
  \end{small}
  \begin{center}     
    \subfloat[Example of wrong alignment]{\includegraphics[width=0.99\columnwidth]{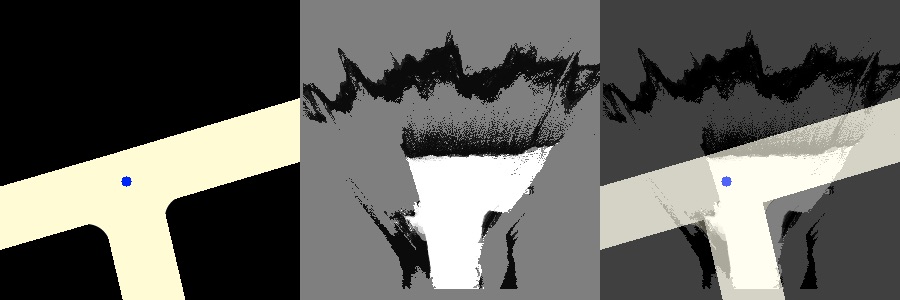}\label{fig:2845_gt_pcr_overlay}}  
  \end{center}
  \caption{For each row, from left to right: the EXPECTEDOG created using the GPS-RTK and the digital map, the SENSORSOG and finally their overlay.
  A misalignment between the GPS-RTK pose (and the associated EXPECTEDOG) and the map is highlighted in the \Cref{fig:2845_gt_pcr_overlay}. 
  It is worth to note that alignment error might come from degraded GPS-RTK measures and/or digital map misalignments.}
  \label{fig:confrontoGT}
\end{figure}

\begin{enumerate}
  \item We select a frame where the vehicle is very near to the intersection;
  \item we compute the SENSOROG, \ie the occupancy grid built using the PC reconstructed from the cameras, see \cite{Ballardini2017};
  \item we generate the EXPECTEDOG, \ie the occupancy grid obtained from OpenStreetMap when hypothesizing the vehicle in the GT position;
  \item we superimpose the two occupancy grids and,
  \item depending on whether the two are aligned, see \Cref{fig:confrontoGT}, we consider the approach to that intersection as usable or not.
\end{enumerate}

Therefore, the localization performances will be benchmarked only in situations where it has been manually verified that the SENSOROGs, which are of course generated from the real positions of the vehicle, are correctly aligned \wrt the corresponding EXPECTEDOGs, computed considering the vehicle in the position GT, \ie a very good estimate of the real position of the vehicle.

\begin{figure}
  \begin{center} \includegraphics[width=.49\textwidth]{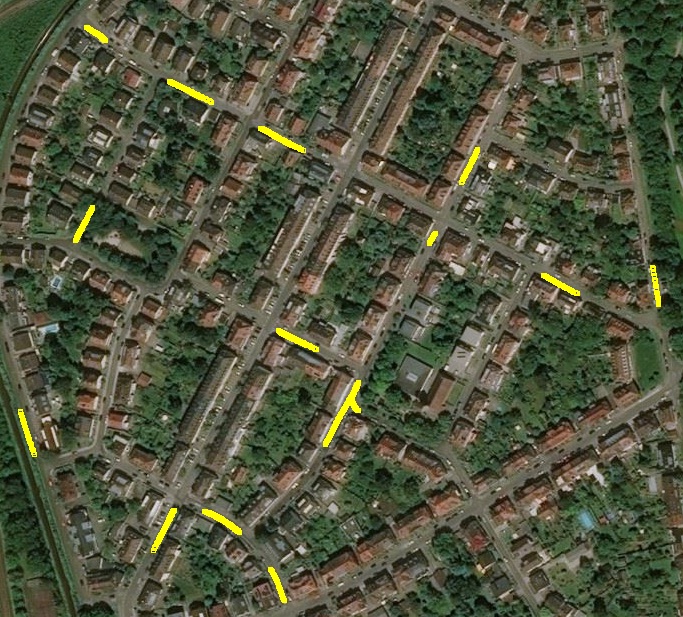} \end{center}
\caption{In the figure the intersections where the 48 usable subsequences have been taken, are highlighted in yellow.}
\label{fig:partibuone}
\end{figure}

In the end we obtained 48 well-aligned subsequences of approaches to different intersection geometries. The involved intersections are shown in \Cref{fig:partibuone}.

\section{Results and Discussion}\label{sec:results}
The analysis of the sequences of the KITTI dataset, presenting subsequences of the vehicle approaching an intersection, is presented in \Cref{tab:tabella}. In many cases the detection of the intersection has not been possible because of the lack of frames, \ie very few frames were recorded during the approach to the intersection, likely because the dataset was recorded at 10Hz. In other cases the light was too low or too much for the cameras to record usable images, or the topology of the intersection cannot be represented with the used intersection model, or the GT could not be verified against aerial imagery. All these cases has been discarded.

\begin{table}
\centering
\begin{threeparttable}
\caption{Results on KITTI residential sequences}
\begin{tabular}{llllll}
dataset name (chars       & Lenght        & frames in              & passed                    & usable            & issues                         \\
(``2011\_'' deleted)      & [mm:ss]       & sequence               & intersect.                & intersect.        &                                \\
09\_26\_drive\_0019       & 00:48         & 487                    & ? 0                       & 0                 & ? -                            \\
09\_26\_drive\_0020       & 00:09         & 92                     & ? 0                       & 0                 & ? -                            \\
09\_26\_drive\_0022       & 01:20         & 806                    & ? 3                       & 0                 & ? -                            \\
09\_26\_drive\_0023       & 00:48         & 480                    & ? 0                       & 0                 & ? -                            \\
09\_26\_drive\_0035       & 00:13         & 137                    & ? 1                       & 1                 & ? -                            \\
09\_26\_drive\_0036       & 01:20         & 809                    & ? 2                       & 0                 & ? {[}1{]}                      \\
09\_26\_drive\_0039       & 00:40         & 401                    & ? 2                       & 2                 & ? -                            \\
09\_26\_drive\_0046       & 00:13         & 131                    & ? 1                       & 1                 & ? -                            \\
09\_26\_drive\_0061       & 01:10         & 709                    & ? 3                       & 0                 & ? {[}1{]}{[}N{]}               \\
09\_26\_drive\_0064       & 00:57         & 576                    & ? 4                       & 4                 & ? {[}N{]}                      \\
09\_26\_drive\_0079       & 00:10         & 107                    & ? 0                       & 0                 & ? -                            \\
09\_26\_drive\_0086       & 01:11         & 711                    & ? 0                       & 0                 & ? -                            \\
09\_26\_drive\_0087       & 01:13         & 735                    & ? 1                       & 0                 & ? -                            \\
09\_30\_drive\_0018       & 04:36         & 2768                   & ? 17                      & 14                & ? -                            \\
09\_30\_drive\_0020       & 01:51         & 1111                   & ? 6                       & 0                 & ? {[}1{]}{[}N{]}               \\
09\_30\_drive\_0027       & 01:51         & 1112                   & ? 7                       & 5                 & ? {[}1{]}                      \\
09\_30\_drive\_0028       & 08:38         & 5183                   & ? 34                      & 0                 & ? {[}1{]}{[}2{]}{[}3{]}{[}N{]} \\
09\_30\_drive\_0033       & 02:40         & 1600                   & ?                         & 0                 & ? {[}1{]}{[}N{]}               \\
09\_30\_drive\_0034       & 02:03         & 1230                   & ? 3                       & 0                 & ? -                            \\
10\_03\_drive\_0027       & 07:35         & 4550                   & ? 41                      & 12                & ? {[}1{]}                      \\
10\_03\_drive\_0034       & 07:46         & 4669                   & ?                         & 9                 & ? -    
\end{tabular}
    \label{tab:tabella}
    \begin{tablenotes}      
      \footnotesize      
      \item
      \begin{enumerate}
       \item [{[1]}] Intersection topology cannot be represented using the proposed intersection model 
       \item [{[2]}] Clear GT misalignments.
       \item [{[3]}] Misalignment OSM \wrt GT or vice-versa.
       \item [{[N]}] Missing aerial image for checking the position GT
      \end{enumerate}      

    \end{tablenotes}
  \end{threeparttable}
\end{table}


\section*{Conclusions}
We analyzed many publicly available datasets in order to build a dataset of sequences for benchmarking methods for vision-based localization at intersection. The sequences have to include streams from a colour stereo rig, altogether with an accurate position estimate to be used as position GT.

We discarded all datasets, apart the residential sequences from the KITTI dataset. The analysis of these sequences led to the selection of 48 subsequences of the vehicle approaching an intersection, usable for the mentioned benchmarking task.

Even though this dataset is a valuable tool, it does not include all types of intersections, not all light conditions, and not all traffic levels. For these reasons it has to be taken as a first step toward the creation of a dataset usable for realistic benchmarking of localization at intersections. With ``realistic'' we mean that the coverage in terms of light conditions, intersection topology, and traffic level allow to consider a method, successful against the benchmark, to be worth trying in worldwide conditions.

A last word of appreciation goes to the KITTI team, whose work turned out to be the only usable dataset, so many years after its collection!

\section*{Appendix: KITTI Residential Sequences}\label{sec:appendixkittiresidentialsequences}
In this Appendix we just present the trajectory of the vehicle in the KITTI residential sequences, to have an overview of the intersections that are potentially available.

\begin{figure}
  \begin{center} \includegraphics[width=0.49\textwidth]{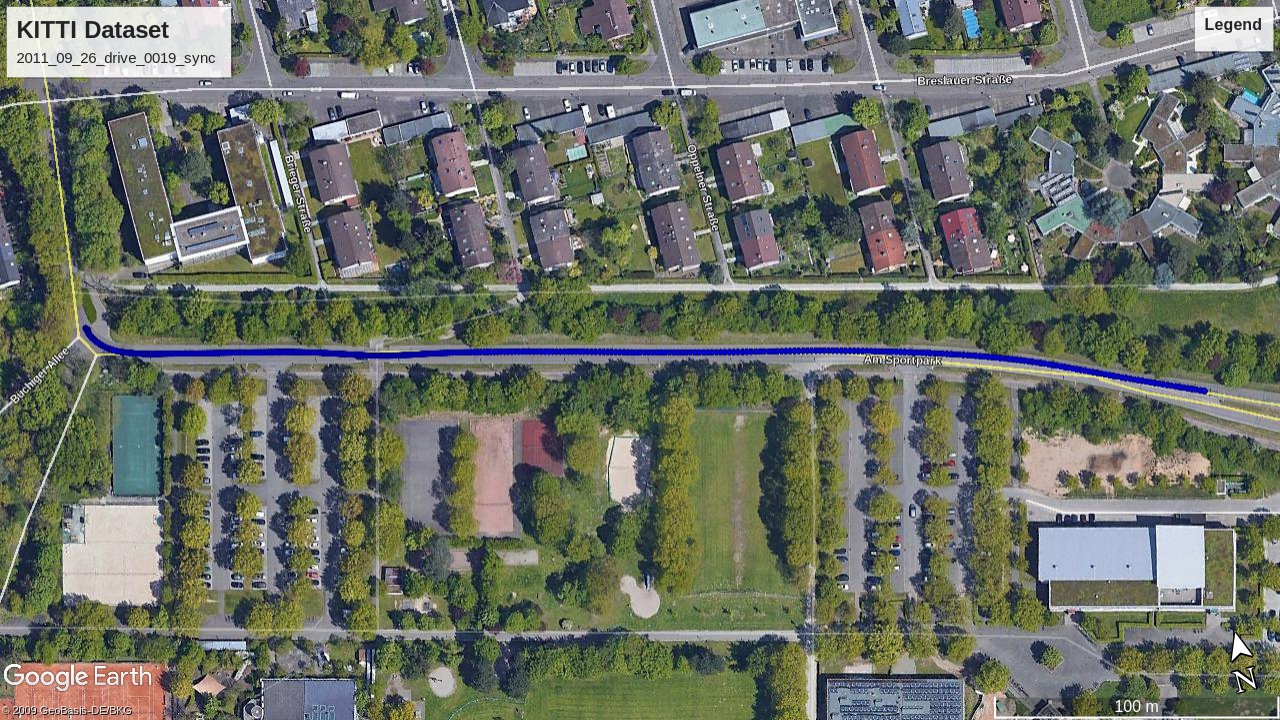} \end{center}
\label{fig:2011_09_26_drive_0019}
\end{figure}

\begin{figure}
  \begin{center} \includegraphics[width=0.49\textwidth]{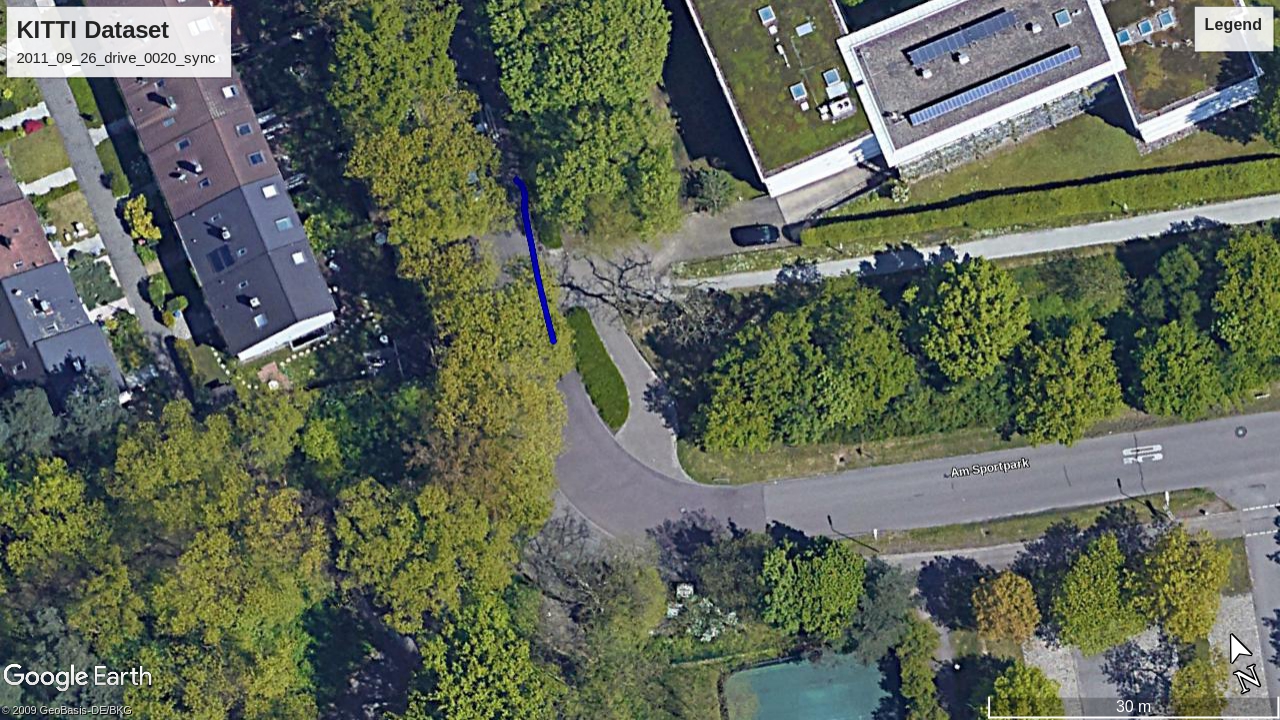} \end{center}
\label{fig:2011_09_26_drive_0020}
\end{figure}

\begin{figure}
  \begin{center} \includegraphics[width=0.49\textwidth]{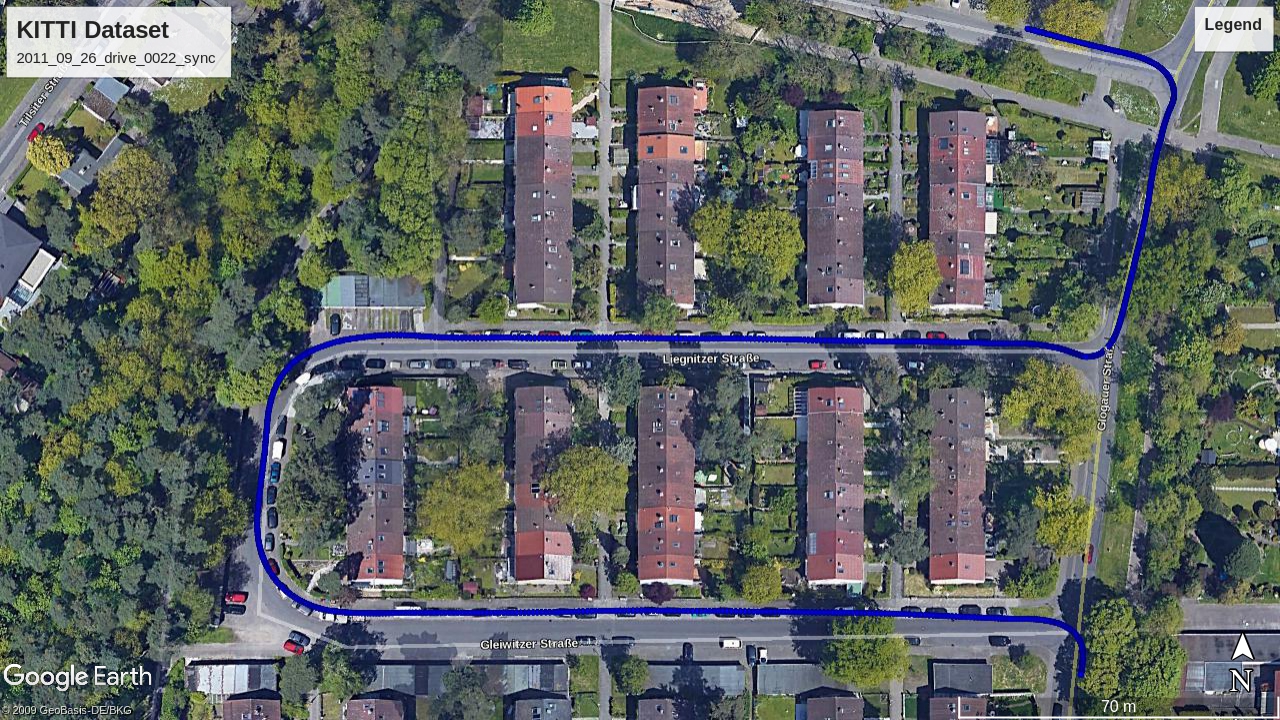} \end{center}
\label{fig:2011_09_26_drive_0022}
\end{figure}

\begin{figure}
  \begin{center} \includegraphics[width=0.49\textwidth]{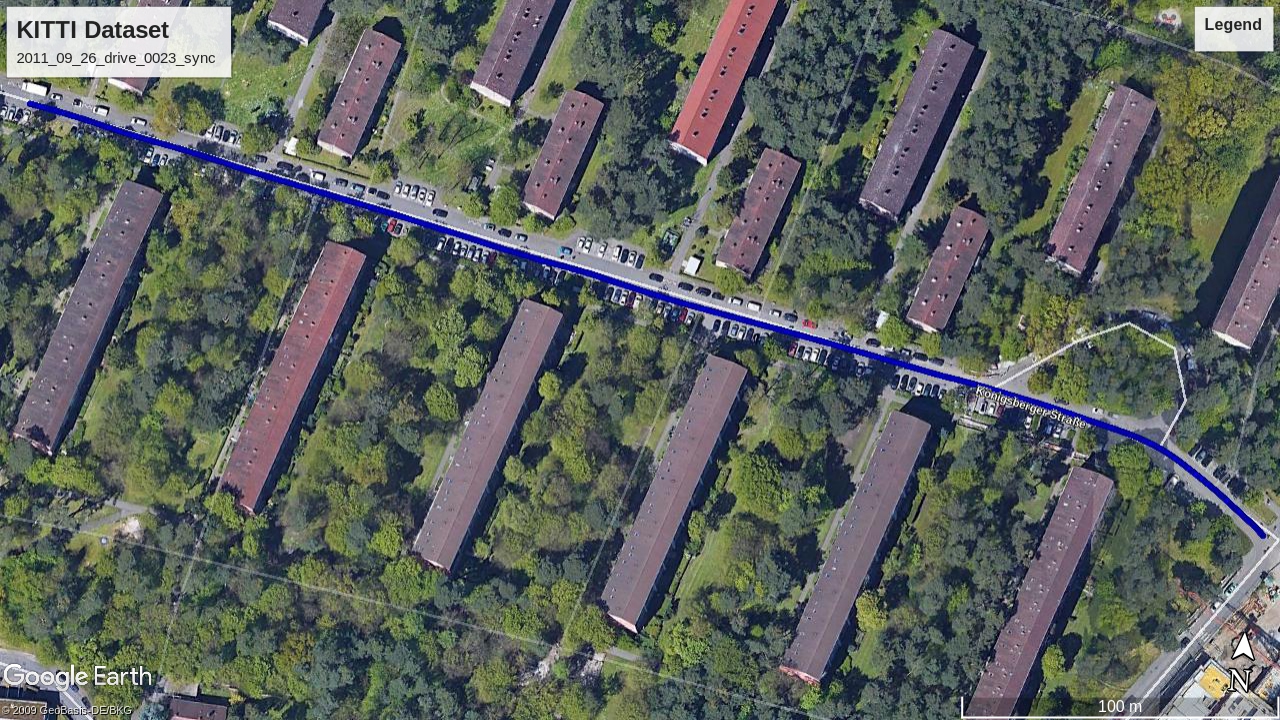} \end{center}
\label{fig:2011_09_26_drive_0023}
\end{figure}

\begin{figure}
  \begin{center} \includegraphics[width=0.49\textwidth]{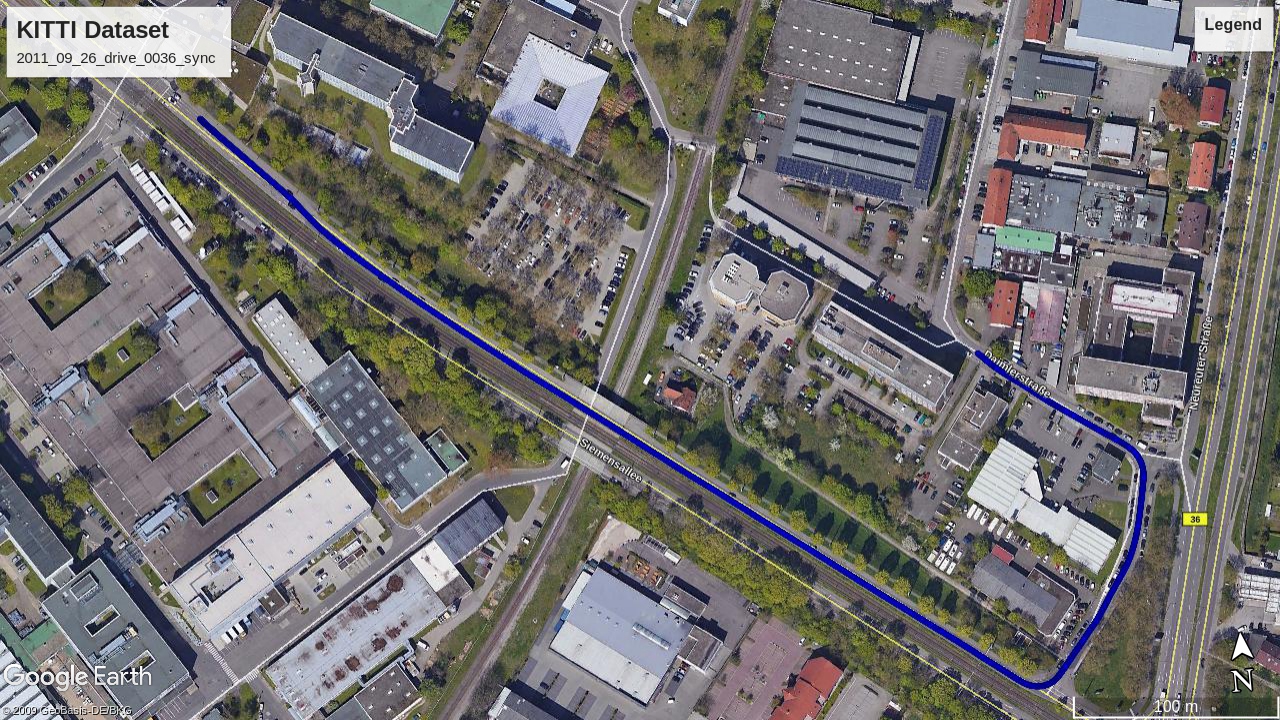} \end{center}
\label{fig:2011_09_26_drive_0036}
\end{figure}

\begin{figure}
  \begin{center} \includegraphics[width=0.49\textwidth]{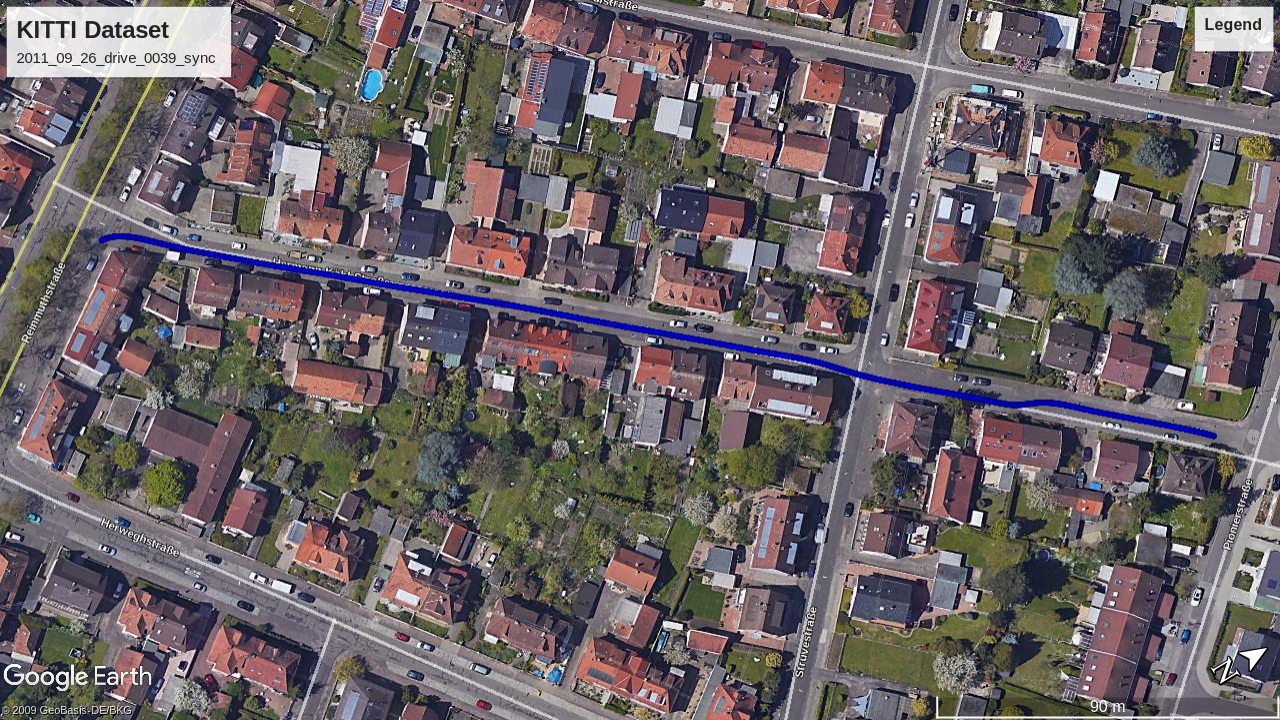} \end{center}
\label{fig:2011_09_26_drive_0039}
\end{figure}

\begin{figure}
  \begin{center} \includegraphics[width=0.49\textwidth]{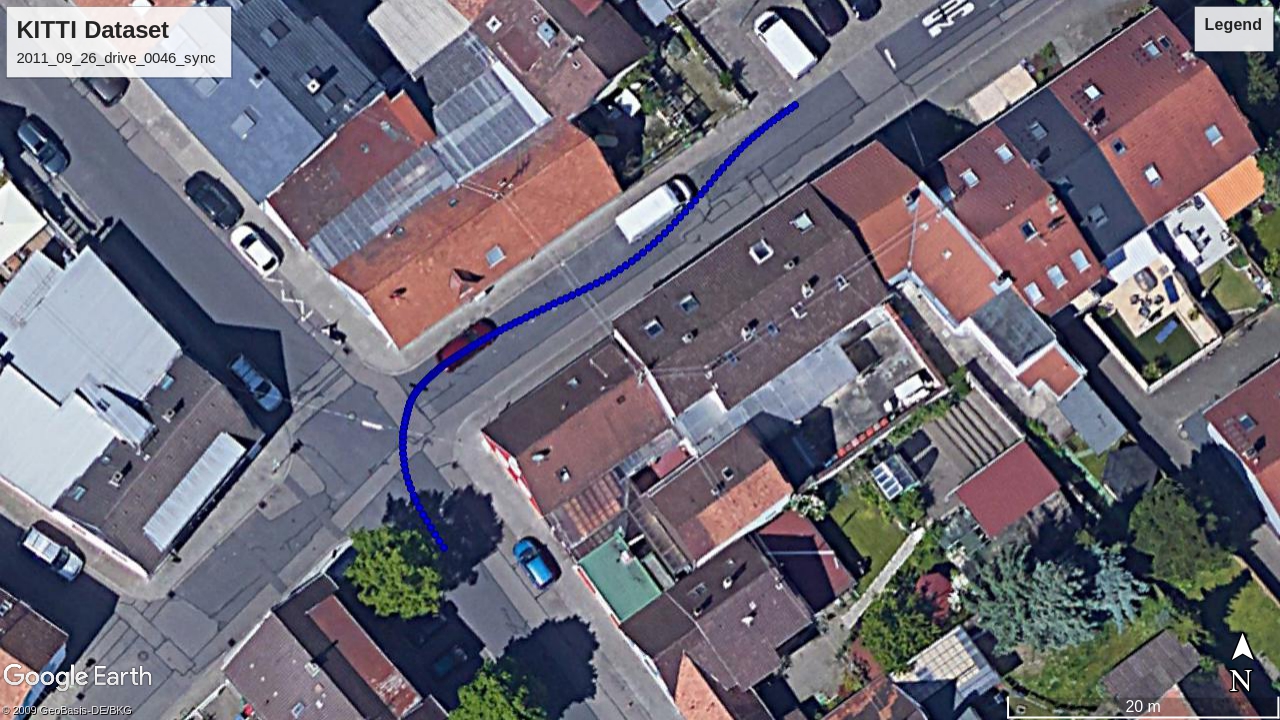} \end{center}
\label{fig:2011_09_26_drive_0046}
\end{figure}

\begin{figure}
  \begin{center} \includegraphics[width=0.49\textwidth]{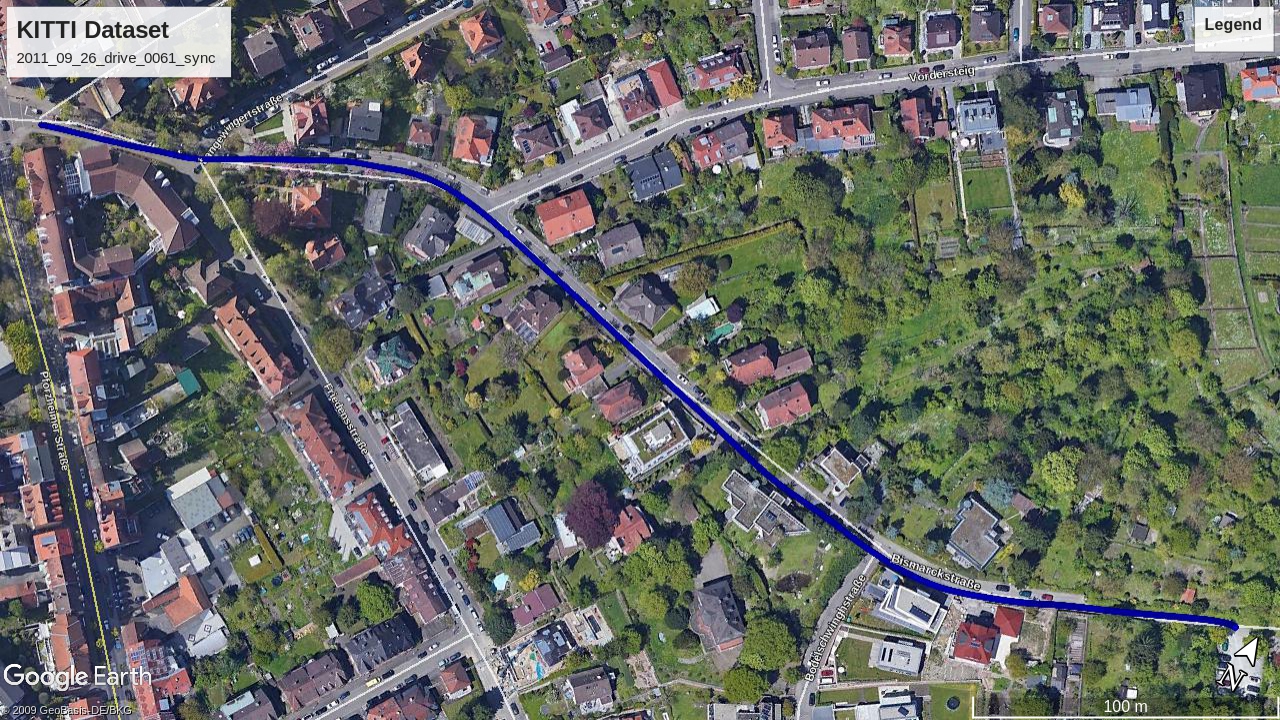} \end{center}
\label{fig:2011_09_26_drive_0061}
\end{figure}

\begin{figure}
  \begin{center} \includegraphics[width=0.49\textwidth]{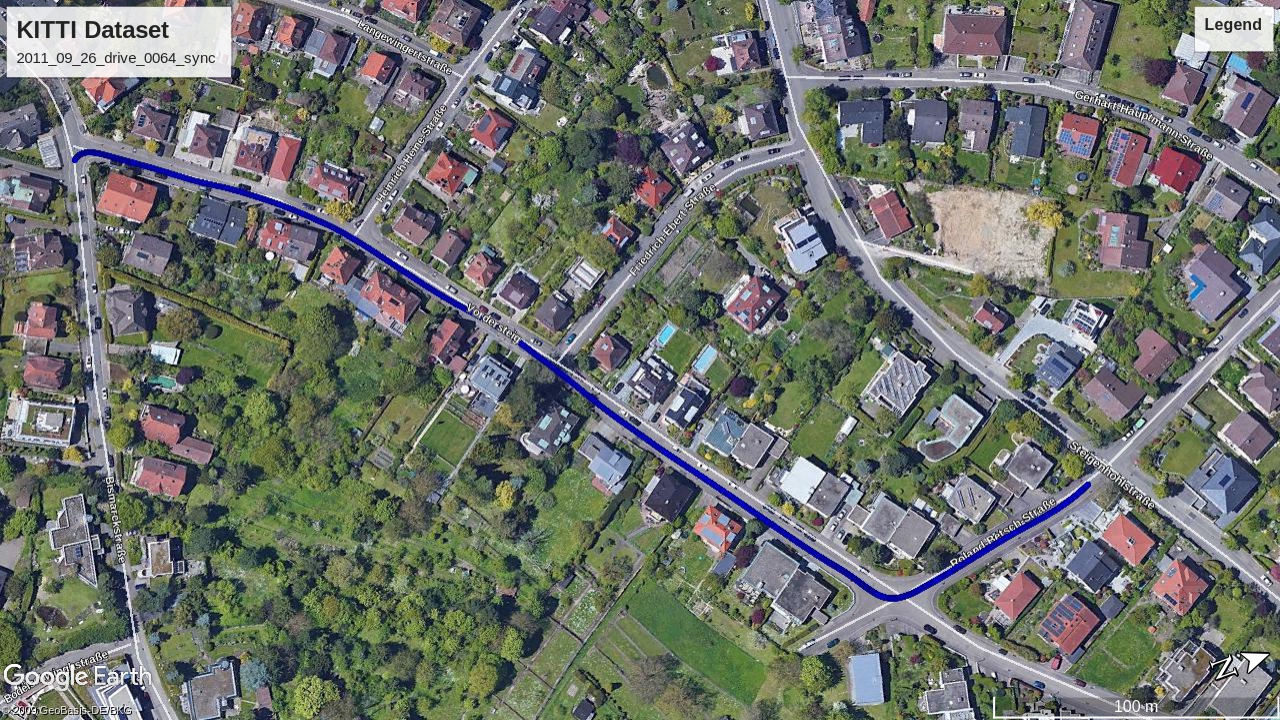} \end{center}
\label{fig:2011_09_26_drive_0064}
\end{figure}

\begin{figure}
  \begin{center} \includegraphics[width=0.49\textwidth]{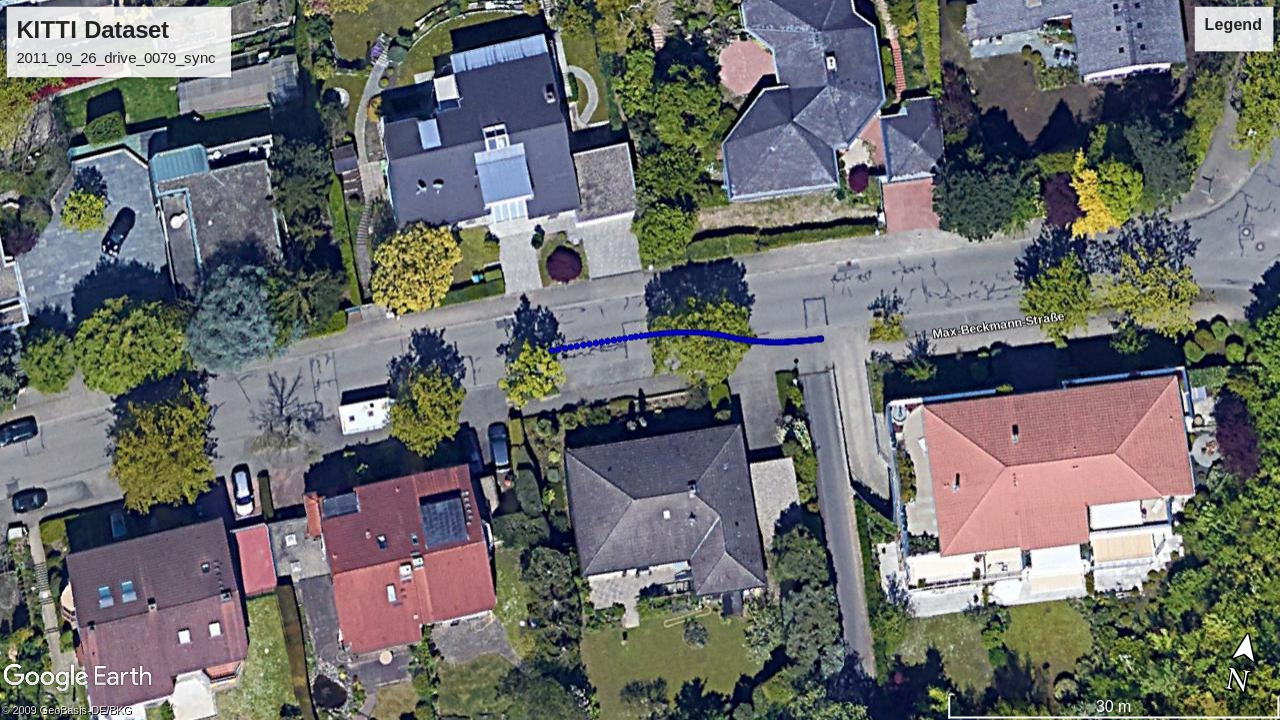} \end{center}
\label{fig:2011_09_26_drive_0079}
\end{figure}

\begin{figure}
  \begin{center} \includegraphics[width=0.49\textwidth]{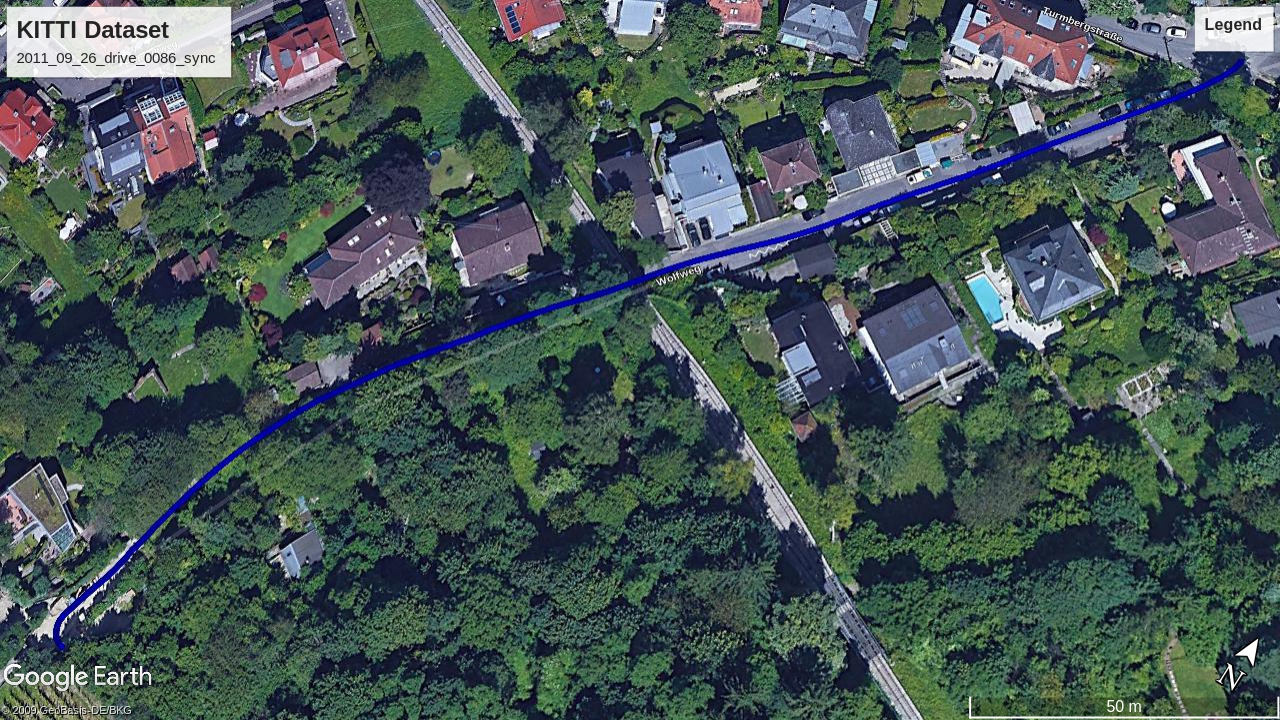} \end{center}
\label{fig:2011_09_26_drive_0086}
\end{figure}

\begin{figure}
  \begin{center} \includegraphics[width=0.49\textwidth]{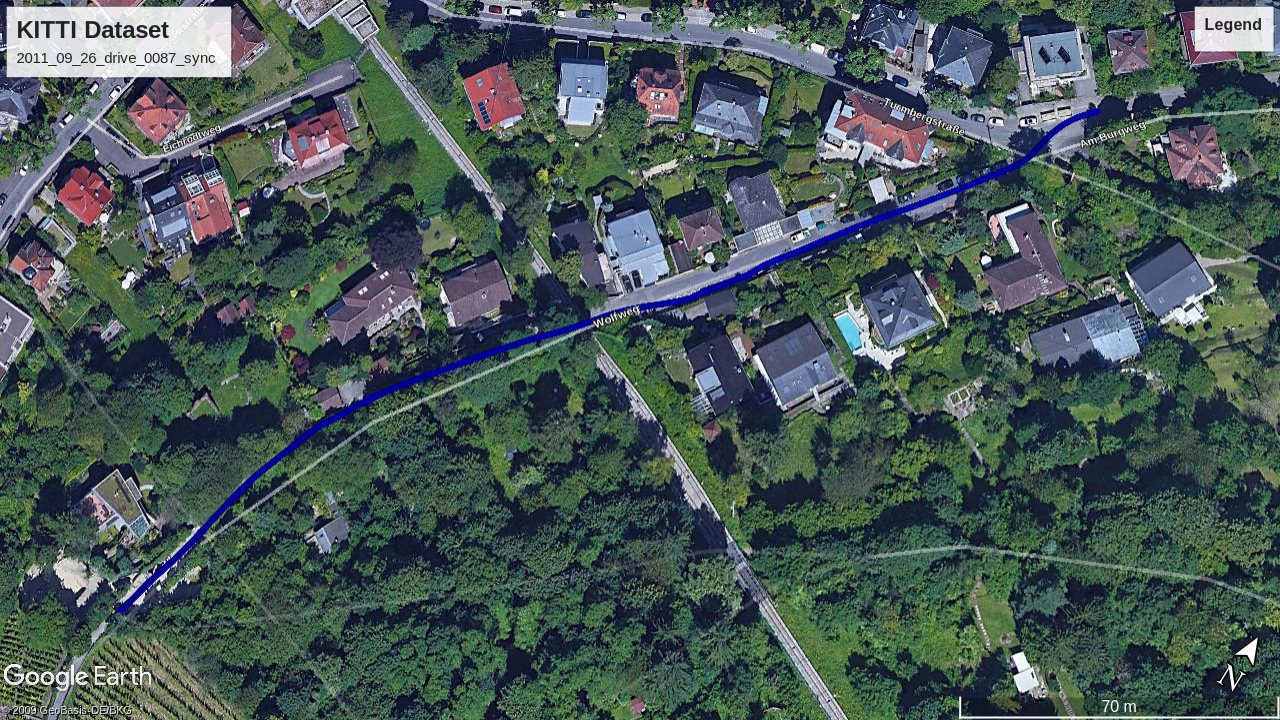} \end{center}
\label{fig:2011_09_26_drive_0087}
\end{figure}

\begin{figure}
  \begin{center} \includegraphics[width=0.49\textwidth]{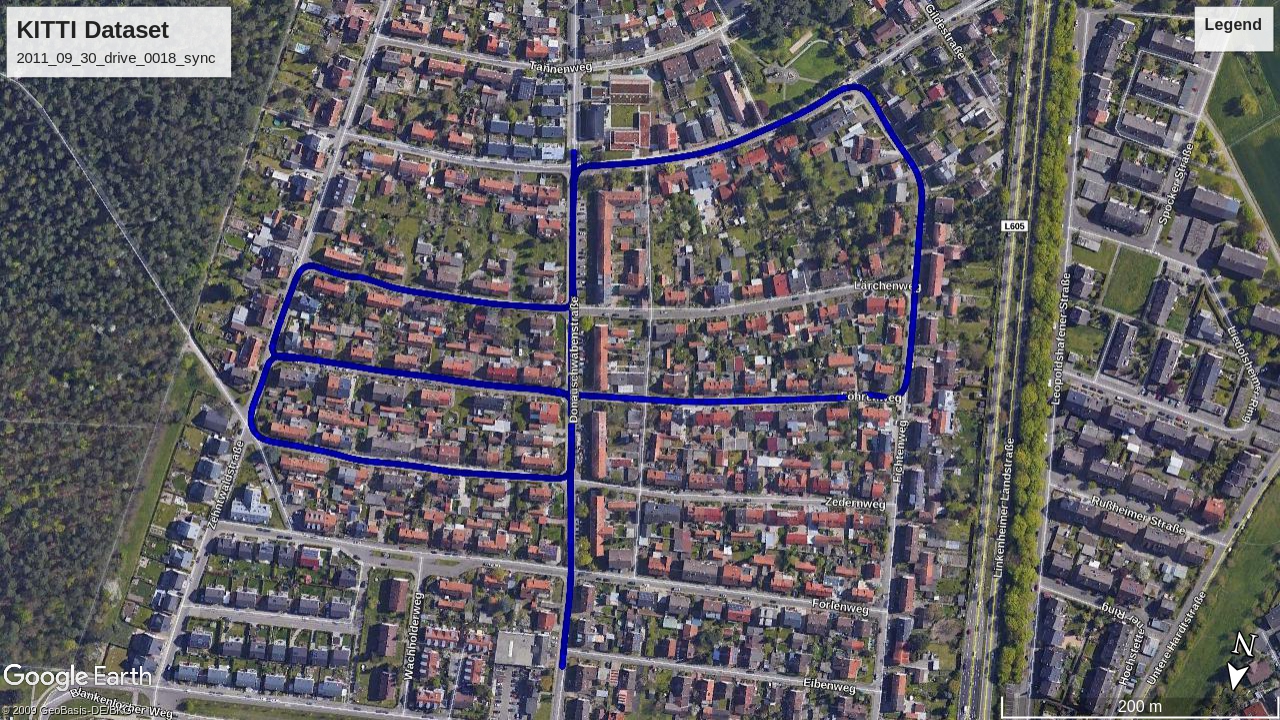} \end{center}
\label{fig:2011_09_30_drive_0018}
\end{figure}

\begin{figure}
  \begin{center} \includegraphics[width=0.49\textwidth]{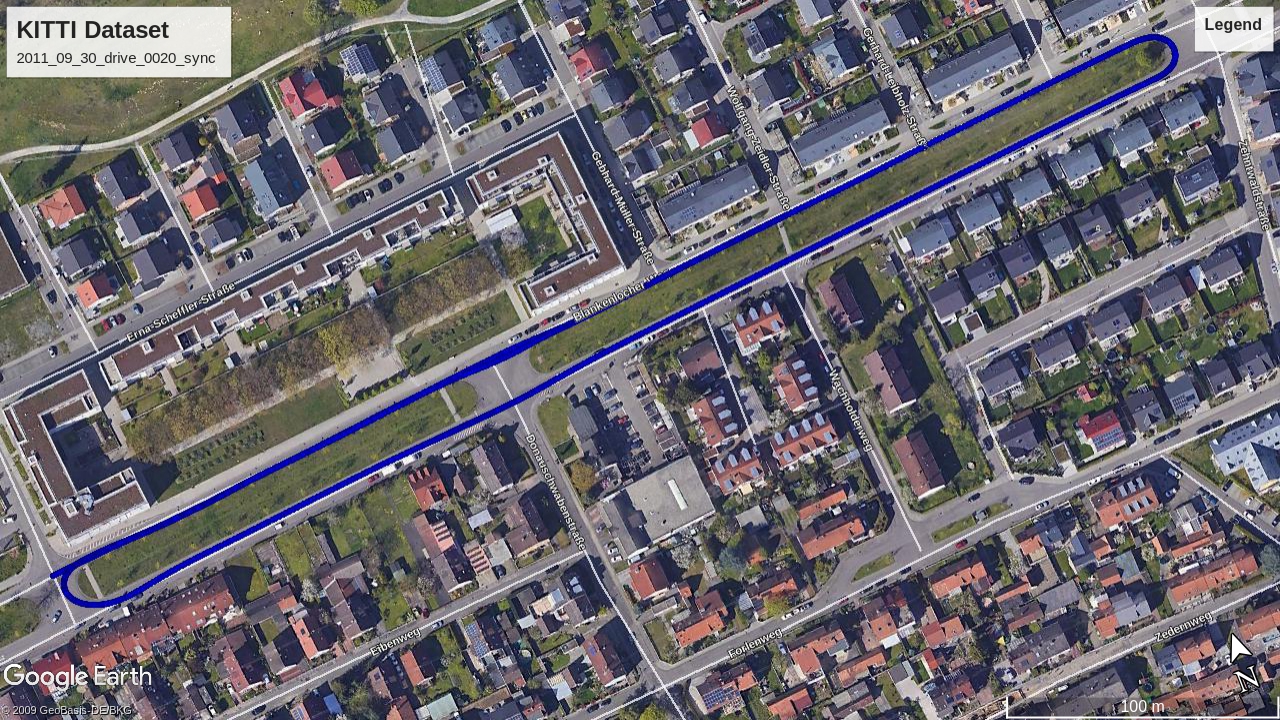} \end{center}
\label{fig:2011_09_30_drive_0020}
\end{figure}

\begin{figure}
  \begin{center} \includegraphics[width=0.49\textwidth]{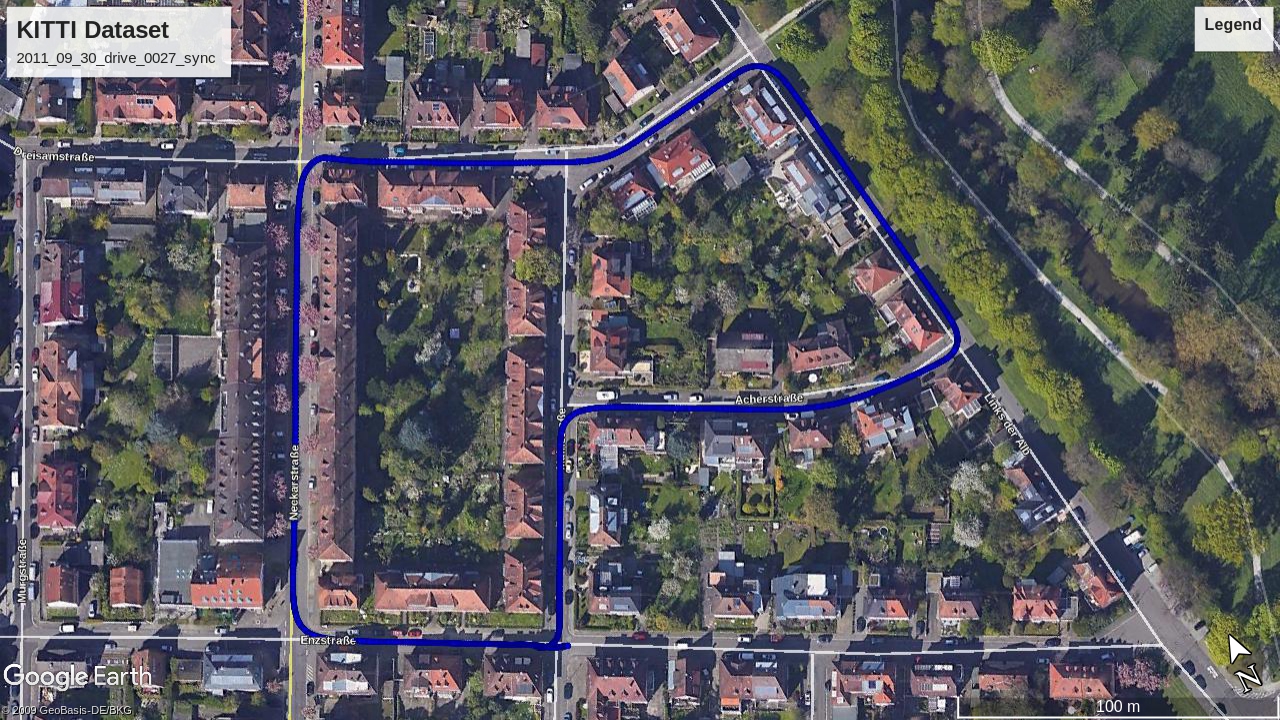} \end{center}
\label{fig:2011_09_30_drive_0027}
\end{figure}

\begin{figure}
  \begin{center} \includegraphics[width=0.49\textwidth]{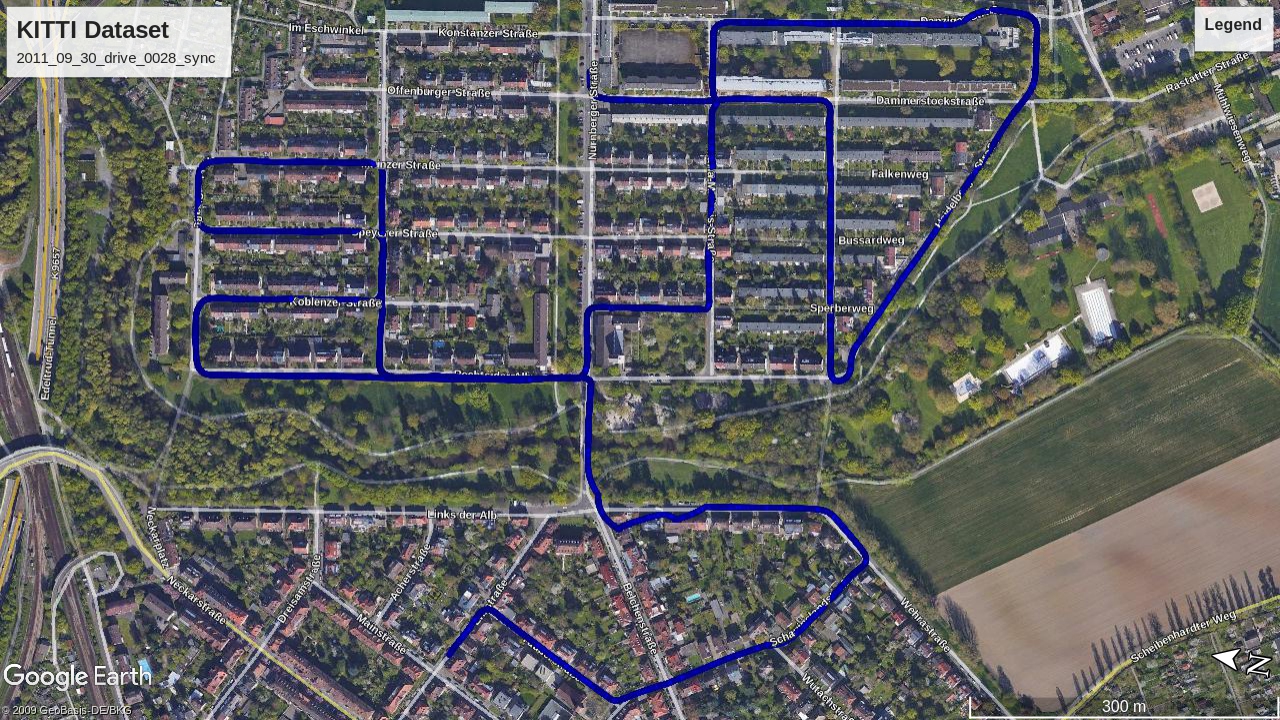} \end{center}
\label{fig:2011_09_30_drive_0028}
\end{figure}

\begin{figure}
  \begin{center} \includegraphics[width=0.49\textwidth]{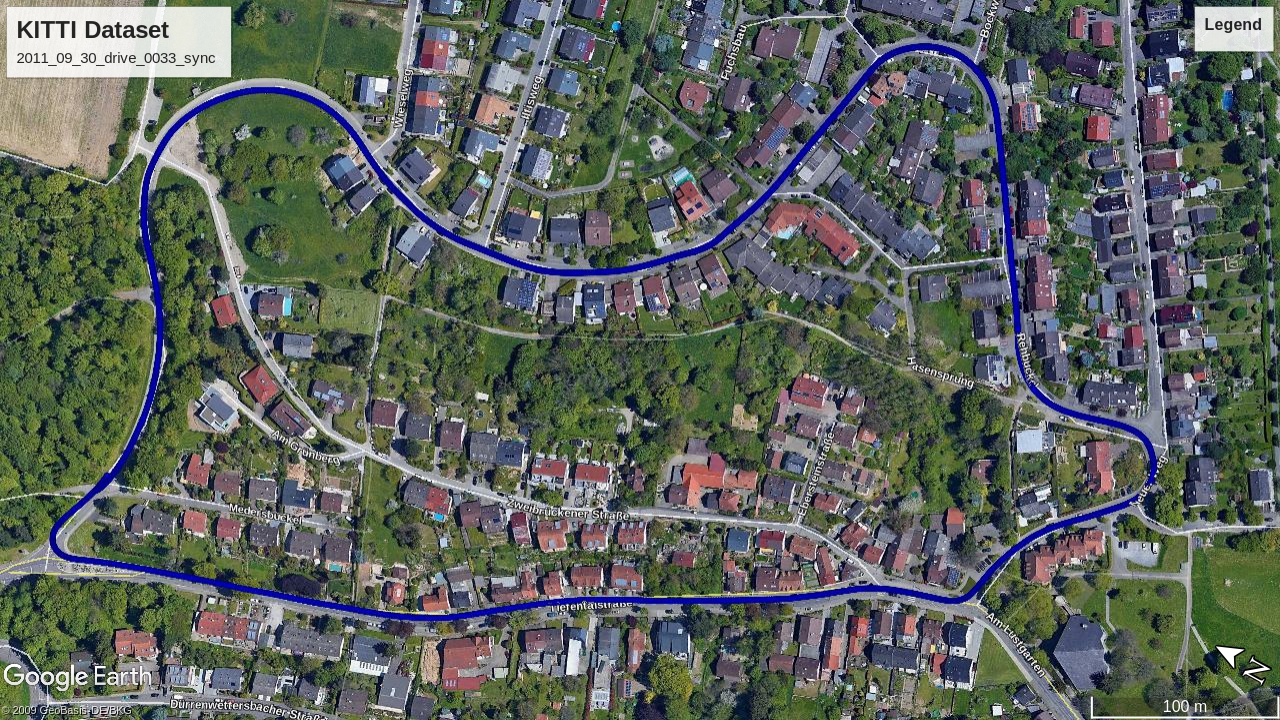} \end{center}
\label{fig:2011_09_30_drive_0033}
\end{figure}

\begin{figure}
  \begin{center} \includegraphics[width=0.49\textwidth]{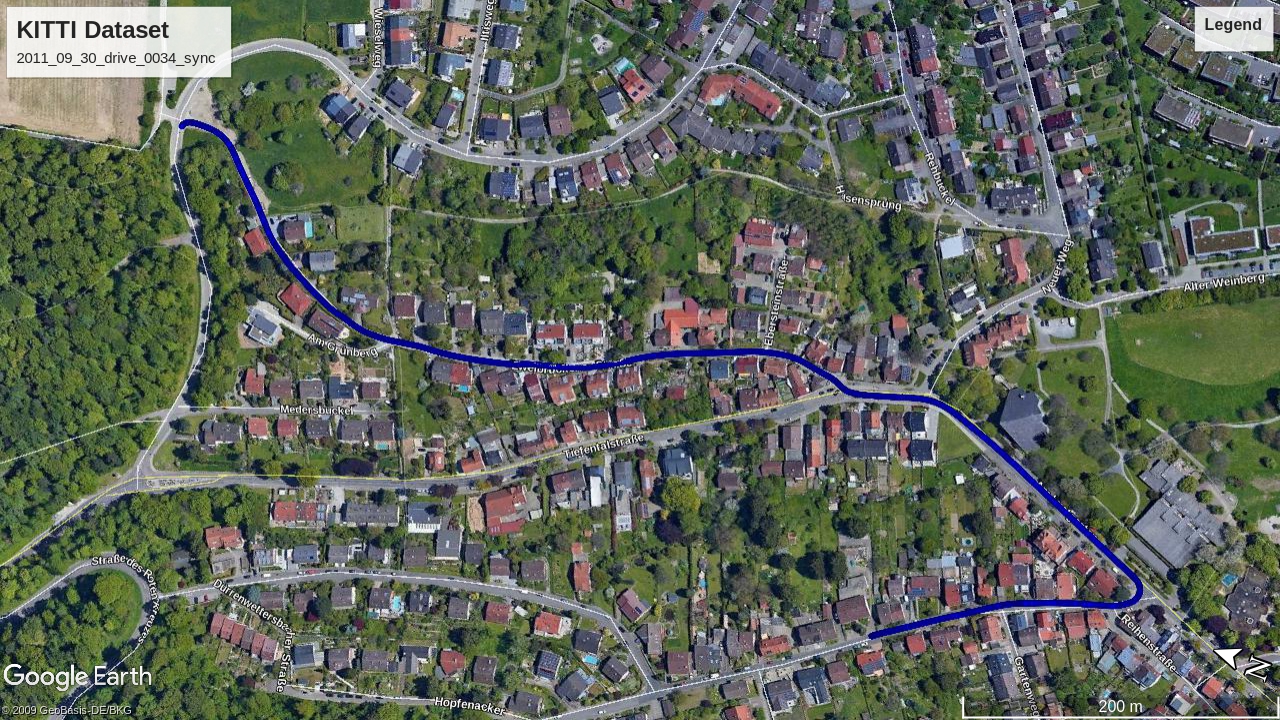} \end{center}
\label{fig:2011_09_30_drive_0034}
\end{figure}

\bibliographystyle{IEEEtran}
\bibliography{IEEEabrv,rle.intersection.dnn}

\end{document}